\documentclass[10pt, conference, compsocconf]{IEEEtran}
\ifCLASSINFOpdf
\else
\fi
\usepackage{graphicx}
\usepackage{amsmath}
\usepackage{amssymb}
\usepackage{subcaption}
\usepackage[mode=buildnew]{standalone}

\begin{document}
\title{Image Compression: Sparse Coding vs. Bottleneck Autoencoders}


\author{Yijing Watkins$^{1,3}$, Oleksandr Iaroshenko$^1$, Mohammad Sayeh$^3$ and Garrett Kenyon$^{1,2}$ \\
   Los Alamos National Laboratory$^1$ \\
  New Mexico Consortium$^2$ \\
  Southern Illinois University Carbondale$^3$  \\
\\
}


%Telephone: (800) 555--1212, Fax: (888) 555--1212}


\maketitle

\begin{abstract}
Bottleneck autoencoders have been actively researched as a solution to image compression tasks. However, we observed that bottleneck autoencoders produce subjectively low quality  reconstructed images.
In this work, we explore the ability of sparse coding to improve reconstructed image quality for the same degree of compression. We observe that sparse image compression produces visually superior reconstructed images and yields higher values of pixel-wise measures of reconstruction quality (PSNR and SSIM) compared to bottleneck autoencoders.
In addition, we find that using alternative metrics that correlate better with human perception, such as feature perceptual loss and the classification accuracy, sparse image compression scores up to 18.06\% and 2.7\%  higher, respectively, compared to bottleneck autoencoders.  Although computationally much more intensive, we find that sparse coding is otherwise superior to bottleneck autoencoders for the same degree of compression.

\end{abstract}


% \fi
\IEEEpeerreviewmaketitle

\section{Introduction}
Image compression methods have significant practical and commercial interest and have been the topic of extensive research. Thumbnail images contain higher frequencies and much less redundancy, which makes them more difficult to compress compared to high-resolution images.
A lot of research has been focused on improving the compression of thumbnails \cite{HinSal06, Krizhevsky09learningmultiple, DBLP:journals/corr/TodericiOHVMBCS15}. 

Bottleneck autoencoders achieve compression by using feed-forward artificial neural networks to reduce the dimensionality of the input data. The basic principles underlying convolutional and fully connected feed-forward neural networks, including bottleneck autoencoders, have been known for years \cite{HinSal06, DeepBoltzmannMachin, Watkins}. 

Sparse coding algorithms
use an overcomplete set of non-orthogonal basis functions (feature vectors) to find a sparse combination of non-zero activation coefficients that most accurately reconstruct each input image. Sparse coding image compression combines sparse coding with the ideas of downsampling and compressive sensing \cite{Tao_Information, Donoho_Compressed, Watkins:2014}: 1) a subset of the original pixels are used as a compressed image, 2) a minimal set of generators that explains the pixels in the compressed image is identified, and 3) the missing pixel values are inferred.

The peak-signal-to-noise ratio (PSNR) and the structural similarity index measure (SSIM) are two commonly used pixel-level image quality metrics. Both PSNR and SSIM fail to capture differences at the feature level and correlate poorly with human perception of image quality. Several researchers have therefore attempted to define alternative measures of compression quality based on the similarity of the features extracted from the reconstructed and original images and have shown that these alternative measures correlate better with human subjective perception \cite{DBLP:conf/eccv/JohnsonAF16, DB16c, SajSchHir17}.   We further introduce a new feature-based measure of compression quality based on loss of classification accuracy, in which compressed images are labeled using  TensorFlow's CIFAR-10 DCNN classifier \cite{tensorflow2015-whitepaper} and the results compared to the baseline performance achieved on the original images.

In this work we apply bottleneck autoencoders and sparse coding approaches to the compression of thumbnail images and use both subjective, pixel-level and feature-based criteria for evaluating reconstructed image quality. We report that sparse coding with random dropout masks produces subjectively superior reconstructed images along with lower feature perceptual loss and higher classification accuracy but yields lower values of pixel-wise metrics such as PSNR and SSIM compared to the bottleneck autoencoders.  However, sparse coding with a checkerboard mask yields superior performance as measured by all three of the above criteria.  

\section{Methods}
\subsection{Sparse Coding}

Given an overcomplete basis, sparse coding algorithms seek to identify the minimal set of generators that most accurately reconstruct each input image. In neural terms, each neuron is a generator that adds its associated feature vector to the reconstructed image with an amplitude equal to its activation. For any particular input image, the optimal sparse representation is given by the vector of neural activations that minimizes both image reconstruction error and the number of neurons with non-zero activity.  Formally, finding a sparse representation involves finding a minimum of the following cost function: 
\begin{equation}
E(\ \overrightarrow{I},\phi,\ \overrightarrow{a})=\min\limits_{ \{\overrightarrow{a}, \, \phi \} } \left[  \, \frac{1}{2}  ||  \overrightarrow{I} - \phi * \overrightarrow{a} ||^2 +	\lambda || \overrightarrow{a} ||_1\right],
\label{eq:SC}
\end{equation}
In Eq.~(\ref{eq:SC}), $\overrightarrow{I}$ is an image unrolled into a vector, and $\phi$ is a dictionary of feature kernels that are convolved with the sparse representation $\overrightarrow{a}$.  The factor $\lambda$ is a tradeoff parameter; larger $\lambda$ values encourage greater sparsity (fewer non-zero coefficients) at the cost of greater reconstruction error. Both the sparse representation $\overrightarrow{a}$ and the dictionary of feature kernels $\phi$ can be determined by a variety of standard optimization methods. 
Our approach to compute a sparse representation for a given input image is based on a convolutional generalization of a rectifying Locally Competitive Algorithm (LCA) \cite{Olshausen08}. Once a sparse representation for a given input image has been found, the basis elements associated with non-zero activation coefficients are adapted according to a local Hebbian learning rule (with a momentum term for faster convergence) that further reduces the remaining reconstruction error. Starting with random basis elements, dictionary learning was performed via Stochastic Gradient Descent (SGD).  This training procedure can learn to factor a complex, high-dimensional natural image into an overcomplete set of basis vectors that capture the high-dimensional correlations in the data. 

\begin{figure}
  \centering
  \begin{subfigure}{0.4\linewidth}
    \includestandalone[width=\linewidth]{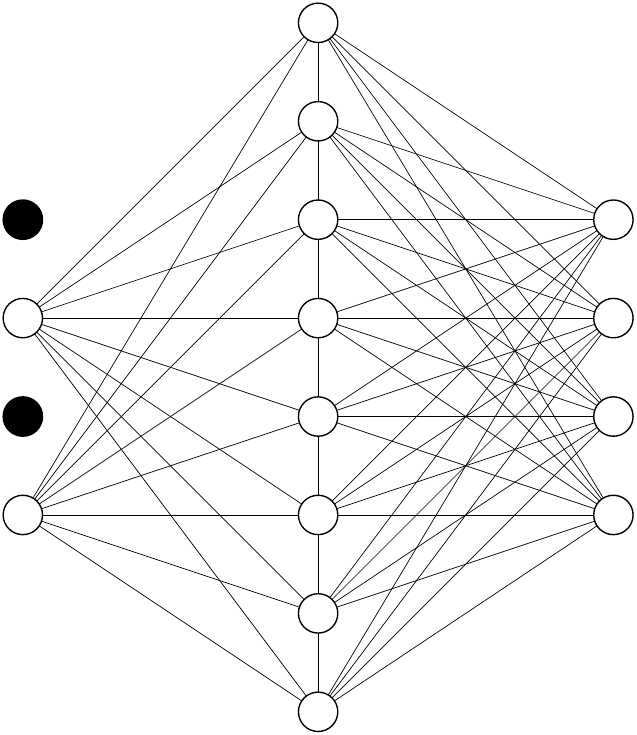}
    \caption{\footnotesize Sparse coding \phantom{second line}}
    \label{fig:sparse}
  \end{subfigure}
  \hspace{0.05\linewidth}
  \begin{subfigure}{0.4\linewidth}
    \includestandalone[width=\linewidth]{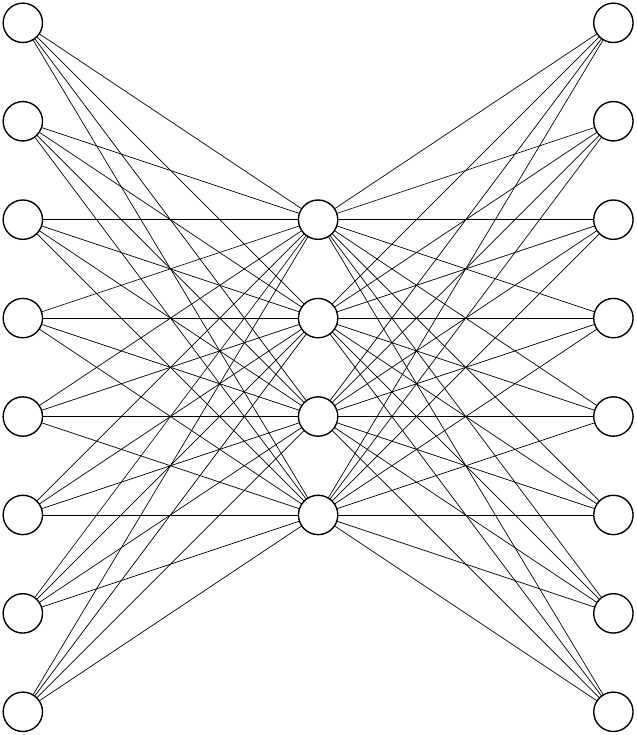}
    \caption{\footnotesize Bottleneck autoencoder}
    \label{fig:bottleneck}
  \end{subfigure}
  \caption{\small (a) Overcomplete representation with pixel dropout (black dots); (b) Undercomplete representation.}
  \label{fig:architectures}
\end{figure}

The sparse coding based image compression architecture is illustrated in Figure~\ref{fig:sparse}, where the input layer (left) is the original image, the hidden layer (middle) is its sparse representation, and the output layer (right) is the image reconstruction.
The white pixels from the input layer are the compressed representation of the original image and black pixels are the omitted pixels of the original image. 
To summarize, image compression based on sparse coding involves two distinct steps:
1) Compression: a subset of the pixels (white circles) is used as a compressed representation of the original image. 
2) Decompression: a minimal subset of pre-trained generators that explains the compressed representation is identified, and the missing pixel values are inferred from these generators.

%
% During image compression, each pixel is dropped with a probability of $50\%$ (black neurons) that creates a random dropout mask.
% %
% The preserved pixels create a compressed

During training, the sparse representation $\overrightarrow{a}$ is found using only the preserved pixel values.  The omitted pixels do not contribute to the minimization of the cost function with respect to  $\overrightarrow{a}$. Then $\overrightarrow{a}$ is used to update the dictionary of generators $\phi$ in order to reconstruct the original image, including the masked pixels, with minimal error.  

We evaluated several sparse coding models which included a variety of  structural elements such as convolutional layers, fully connected layers and multi-scale dictionaries.   To facilitate comparison, each model contained the same total number of neurons.
% such as sparse convolutional models, multi-scaled fully connected and convolutional models, and hierarchical multi-scaled fully connected and convolutional with top-down feedback models.
We found that the best performance was achieved by a model consisting of a single convolutional layer containing 1024  features with a patch size of $16\times16$ pixels and a stride of 2 ($\simeq42.6$ times overcomplete).

\begin{figure}
  \centering
\includegraphics[height=3.5cm]{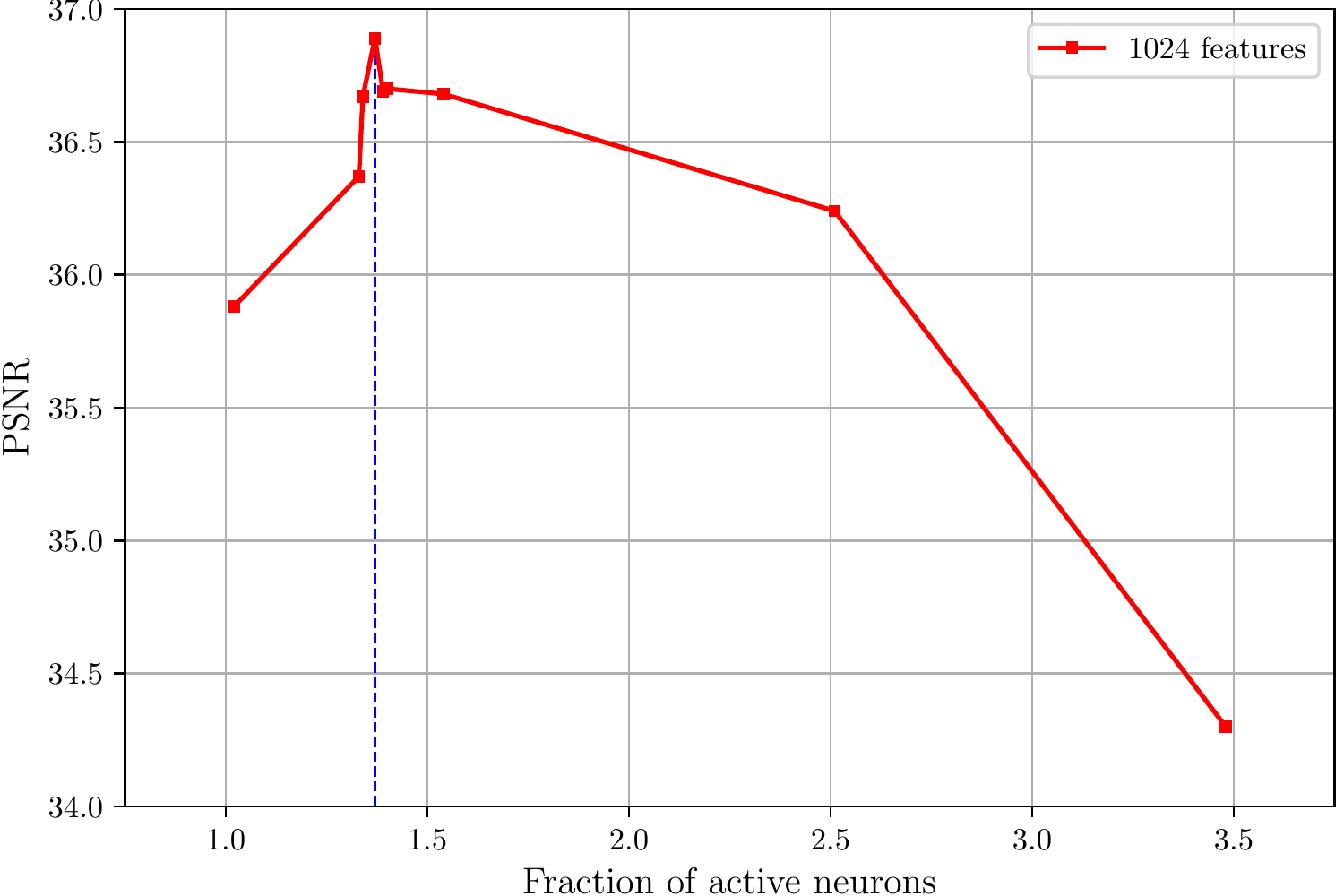}
  \caption{\small PSNR as a function of sparsity (fraction of active neurons) for various values of the tradeoff parameter $\lambda$.}
  \label{fig:sparisity}
\end{figure}

To determine the optimum percentage of non-zero activity, or optimal tradeoff parameter $\lambda$, for our sparse coding model, we trained our network for one epoch on 9 different $\lambda$ values using the single layer sparse convolutional model. We found that for the optimum value of $\lambda$ for our network, the average percentage of active neurons is 1.37\% (see Figure \ref{fig:sparisity}). Our sparse coding image compression results used this single sparse convolutional model with the aforementioned optimal level of sparsity. 

% \begin{figure}[t]
%   \centering
% %   \includegraphics[width=0.6\linewidth]{sparse_coding.png}
%   \includestandalone[width=0.6\linewidth]{fig/sparse}
%   \caption{Sparse Coding Architecture: Overcomplete representation with random pixel dropout (black dots)}
%   \label{fig:sparse}
% \end{figure}

% \begin{figure}[t]
%   \centering
%   \includestandalone[width=0.6\linewidth]{fig/bottleneck}
%   \caption{Bottleneck Auto-Encoder Architecture: Undercomplete representation}
%   \label{fig:bottleneck}
% \end{figure}

\subsection{Bottleneck Autoencoder}

The bottleneck autoencoder architecture is illustrated in Figure~\ref{fig:bottleneck}. This structure includes one input layer (left), one or more hidden layers (middle), and one output layer (right). The input and output layers contain the same number of neurons, where the output layer aims to reconstruct the input. Image compression is achieved by restricting the number of neurons in the smallest hidden layer (the bottleneck) to contain half the number of neurons of the input layer to achieve a 2:1 compression ratio. We evaluated several bottleneck autoencoder models such as fully connected models, convolutional models, and multi-scaled models, and found that a model combining one convolutional layer with a stride of 4 and one fully connected layer, each containing the same number of neurons ($1/4$ the number of pixels in the thumbnails), provides higher reconstructed image quality over other bottleneck models. The bottleneck autoencoder network finds the optimal hidden layer features across the training set of input images, that minimize the image reconstruction error. In the simplest case, as illustrated in Figure~\ref{fig:bottleneck}, the model minimizes the following energy function: 
\begin{equation}
\begin{split}
E(x;W,b,W',b') & = ||x - x'||^2 \\
 & = || x - \sigma(W'(\sigma(Wx + b)) + b') ||^2 
\end{split}
\label{eq:BA}
\end{equation}
In Eq.~(\ref{eq:BA}), $x$ is the input image, $x'$ is the reconstructed image, $\sigma$ is the ReLU activation function. $W$ and $b$ are the encoder's weights matrix and a bias vector, $W'$ and $b'$ are the decoder's weights matrix and a bias respectively. 
%In our model, the encoder and decoder share the same weights and bias. In this type of neural network architecture, the structured bottleneck layer could be treated as a nonlinear mapping of input features and the decompressor reconstructs the compressed image back to the neurons in the output layer using the same weights as the compressor. The bottleneck autoencoder model employs SGD with momentum to train optimal values of the weights and bias after being randomly initialized. The bottleneck autoencoder is designed to preserve only those features that best describe the original image and  shed redundant information. 

In this type of neural network architecture, a large number of input neurons is fed into a smaller number of neurons in the hidden layer, which works as the compressor. This structured bottleneck layer could be treated as a nonlinear mapping of input features. The decompressor then reconstructs the compressed image back to the neurons in the output layer using the same weights as the compressor. The bottleneck autoencoder model employs SGD with momentum to train optimal values of the weights and bias after being randomly initialized. The bottleneck autoencoder is designed to preserve only those features that best describe the original image and  shed redundant information. 

\section{Results}
\label{headings}
In this work, we compared bottleneck autoencoders with two sparse coding approaches. For sparse coding we masked $50\%$ of the pixels either randomly or arranged in a checkerboard pattern to achieve a 2:1 compression ratio. 
The random mask is regenerated for every image batch whereas the checkerboard mask is fixed. In order to evaluate and compare the quality of the proposed sparse coding and bottleneck autoencoder image compression models, we used PetaVision\cite{petavision}, an open source neural simulation toolbox that enables multi-node, multi-core and GPU accelerated high-performance implementations of sparse solvers derived from LCA as well as conventional neural network models.
We use the CIFAR-10 dataset, which consists of 50,000 $32\times32$ images for training and 10,000 $32\times32$ images for testing.  Category labels were not used in this study.  

\subsection{Image Compression}
\begin{figure*}
  \centering
  \begin{subfigure}{0.15\linewidth}
    \includegraphics[width=\linewidth]{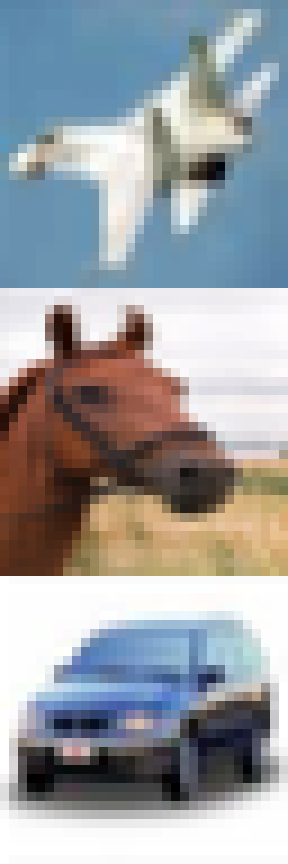}
    \caption{}
    \label{fig:original}
  \end{subfigure}
%   \hspace{0.0\linewidth}
  \begin{subfigure}{0.15\linewidth}
    \includegraphics[width=\linewidth]{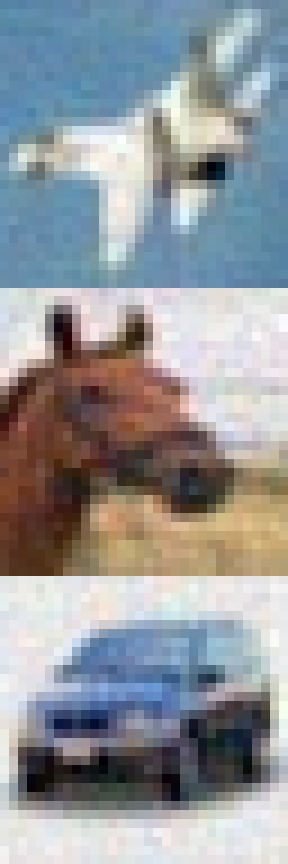}
    \caption{}
    \label{fig:checkerboard-recon}
  \end{subfigure}
%   \hspace{0.0\linewidth}
  \begin{subfigure}{0.15\linewidth}
    \includegraphics[width=\linewidth]{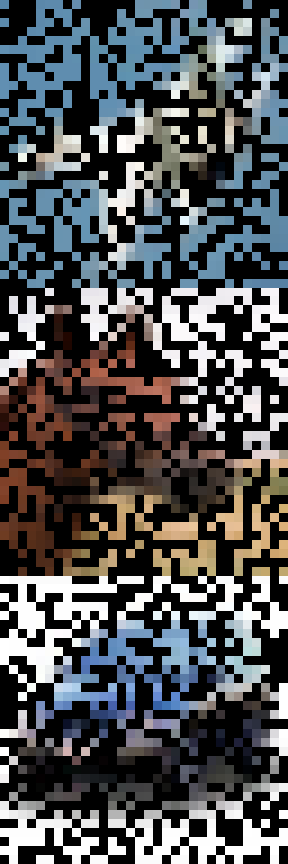}
    \caption{}
    \label{fig:random-recon}
  \end{subfigure}
%   \hspace{0.0\linewidth}
  \begin{subfigure}{0.15\linewidth}
    \includegraphics[width=\linewidth]{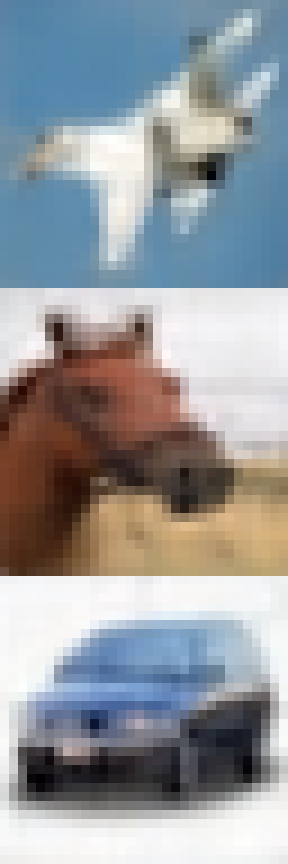}
    \caption{}
    \label{fig:bottleneck-recon}
  \end{subfigure}
%   \hspace{0.0\linewidth}
  \begin{subfigure}{0.15\linewidth}
    \includegraphics[width=\linewidth]{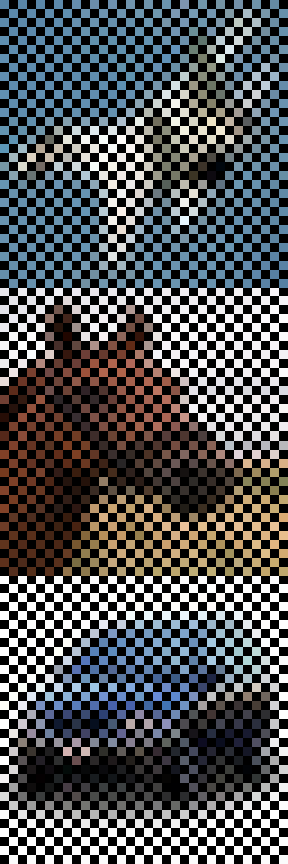}
    \caption{}
    \label{fig:error-checkerboard}
  \end{subfigure}
%   \hspace{0.0\linewidth}
  \begin{subfigure}{0.15\linewidth}
    \includegraphics[width=\linewidth]{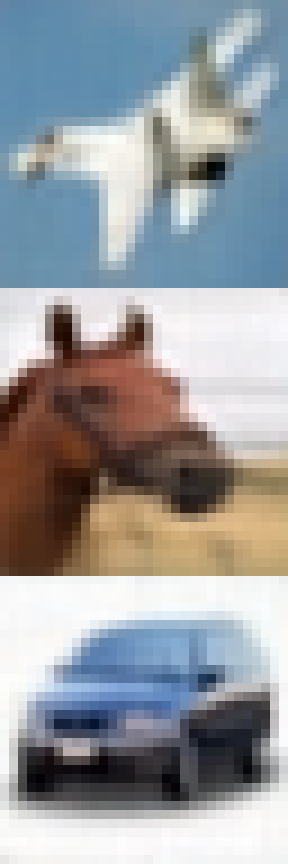}
    \caption{}
    \label{fig:error-random}
  \end{subfigure}
%   \hspace{0.0\linewidth}
  \caption{\small (a) Original images; (b) Reconstructed images: Bottleneck autoencoder; (c) Random masks in which ${\simeq}$50\% of the original pixels are omitted; (d) Reconstructed images: Sparse coding with random mask; (e)  Checkerboard masks in which 50\% of the original pixels are omitted; (f) Reconstructed images: Sparse coding with checkerboard mask.}
  \label{fig:recon}
\end{figure*}

We first evaluated the quality of the reconstructed images from sparse coding and the bottleneck autoencoder using subjective human perceptual judgments. 
Figure \ref{fig:recon} shows examples of sparse coding and bottleneck autoencoder based image compression. 
% Figure \ref{fig:original} shows example images drawn from the original CIFAR-10 testing set, 
% %
% Figure \ref{fig:checkerboard-recon} shows the sparse coding with checkerboard mask reconstructions (50\% compression).
% %
% Figure \ref{fig:random-recon} shows reconstructions from sparse coding using a random mask with the same image compression ratio.
% %
% Figure \ref{fig:bottleneck-recon} shows the bottleneck autoencoder reconstructions when the same image compression ratio is applied.
% %
% Finally, Figure \ref{fig:error-checkerboard} \ref{fig:error-random} \ref{fig:error-bottleneck} show the reconstruction errors for the three methods.
%
Subjective examination reveals that the reconstructed images from sparse coding with either random or checkerboard mask exhibit less noise, has a smoother background, and results in more natural looking reconstructions than images reconstructed from the bottleneck autoencoder.
Sparse coding with a random mask preserves less fine detail compared to the bottleneck autoencoder  whereas sparse coding with a checkerboard mask preserves both background and fine details much better than both the other methods.
Overall, the images compressed using sparse coding with a checkerboard mask are almost visually indistinguishable from the original images.
\subsection{Pixel-wise Loss in Image Space}

We used two well-known pixel-wise image quality metrics: the peak-signal-to-noise ratio (PSNR) and the structural similarity index measure (SSIM) to evaluate the reconstructed images from either sparse coding or the bottleneck autoencoder.
\begin{table}
\caption{Image Reconstruction }
\label{Image Recon}
\centering
\begin{tabular}{lll}
\hline
Methods     & PSNR & SSIM \\
\hline
Bottleneck Autoencoder & 29.107 & 0.905    \\
Sparse Coding with Random Mask & 26.438 & 0.865      \\
Sparse Coding with Checkerboard Mask & 30.090 & 0.940      \\
\hline
  \end{tabular}
\end{table}
Table \ref{Image Recon} indicates that the reconstructed images from sparse coding with random mask contain lower values of PSNR and SSIM compared to the bottleneck autoencoder whereas the corresponding values for the checkerboard masks are higher than for the other methods.  
However, PSNR and SSIM measurements are well known to correlate poorly with human perception of image quality. 

\subsection{Perceptual Loss in Feature Space}

To test the hypothesis that reconstructed images obtained from sparse coding include more information relevant to human perception compared to bottleneck autoencoders, we calculate the feature perceptual loss, given as the Euclidean distance of feature representations between original and reconstructed images\cite{DBLP:conf/eccv/JohnsonAF16, DB16c, SajSchHir17}. To capture and compare the feature representations, we first pre-trained the original (non-compressed) CIFAR-10 training image set on the DCNN classifier from TensorFlow \cite{tensorflow2015-whitepaper}.
%For the feature representations, we used the activations of the second convolutional layer (9216 neurons) and second pooling layer (2304 neurons) when processing the original and the reconstructed images.
Table~\ref{FeaturePerceptualLosses} illustrates Loss~1 and Loss~2, which represent the feature perceptual losses captured from the activations of the second convolutional layer and the second pool layer in the DCNN, respectively.
Overall, the reconstructed images from sparse coding with checkerboard mask and random mask contain on average 18.06\% and 3.74\% lower feature perceptual loss compared to the bottleneck autoencoder.
\begin{table}
\caption{Feature Perceptual Losses}
\label{FeaturePerceptualLosses}
\centering
\begin{tabular}{lclclc}
\hline
Methods     & Loss 1 & Loss 2 \\
\hline
Bottleneck Autoencoder  &    42109  & 36617  \\
Sparse Coding with Random Mask &    39154   & 36446  \\
Sparse Coding with Checkerboard Mask &    29157   & 34653  \\
\hline
\end{tabular}
\end{table}
\subsection{Classification}

To further compare the different compression methods, we checked the classification accuracy of the reconstructed images using the same DCNN classifier.
After three training and testing runs with different random seeds we found that the sparse coding with checkerboard and random masks supported on average 2.7\% and 1.6\% higher classification accuracy compared to the bottleneck autoencoder (see Table~\ref{classification}).
\begin{table}
\caption{Image Classification}
\label{classification}
\centering
\begin{tabular}{lclclc}
\hline
Methods     & Accuracy (\%) \\
\hline
Bottleneck Autoencoder  &     76.0  \\
Sparse Coding with Random Mask &   77.6  \\
Sparse Coding with Checkerboard Mask &    78.7  \\
\hline
\end{tabular}
\end{table}
\section{Conclusion}
Sparse image compression with checkerboard and random masks provides subjectively superior visual quality of reconstructed images, on average 2.7\% and 1.6\% higher classification accuracy and 18.06\% and 3.74\% lower feature perceptual loss, respectively, compared to bottleneck autoencoders.  This paper provides support for the hypothesis that reconstructed images obtained from sparse coding with checkerboard and random masks include more content-relevant information compared to bottleneck autoencoders for the same image compression ratio.  
%whereas sparse coding with random mask yield lower values of pixel-wise metrics such as PSNR and SSIM compared to the bottleneck autoencoder although the corresponding values for the checkerboard masks are higher.  

\section*{Acknowledgment}

This work was funded by the NSF, the DARPA UPSIDE program and the LDRD program at LANL.



%\bibliographystyle{IEEEtran}

% \bibitem{Comp_hinton} Geoffrey Hinton\ \& Ruslan Salakhutdinov, (2006). Reducing the Dimensionality of Data with Neural Networks.  {\it Science}, 313, 504-507. doi: 10.1126/science.1127647 
% \bibitem{cifar_alex} Alex Krizhevsky, (2009). Learning multiple layers of features from tiny images. Technical Report.
% \end{thebibliography}

{\small
\bibliographystyle{ieee}
\bibliography{egbib}
}

\end{document}